\documentclass[nopreprintline,final,12pt,numbers,authoryear]{elsarticle}

\usepackage{amssymb}
\usepackage{graphicx}
\usepackage{color}
\usepackage{rotating}
\usepackage{amsfonts}

\usepackage{amsmath}
\usepackage{algorithm}
\usepackage{algpseudocode}
\usepackage{tabularx}
\usepackage{hyperref}

\newcommand{\tcb}[1]{\textcolor{black}{#1}}

\journal{Expert Systems with Applications}

\begin{document}

\begin{frontmatter}



\title{ODTE - An ensemble of multi-class SVM-based oblique decision trees}


\author[inst1]{Ricardo Monta\~nana}

\affiliation[inst1]{organization={Departamento de Sistemas Informáticos.},
            addressline={Universidad de Castilla-La Mancha}, 
            city={Albacete},
            postcode={02071}, 
            country={Spain}}


\author[inst1]{Jos\'e A. G\'amez}\ead{jose.gamez@uclm.es}
\author[inst1]{Jos\'e M. Puerta}\ead{jose.puerta@uclm.es}

\cortext[cor1]{Corresponding author: Jose A. Gámez}

\begin{abstract}
We propose ODTE, a new ensemble that uses oblique decision trees as base classifiers. Additionally, we introduce STree, the base algorithm for growing oblique decision trees, which leverages support vector machines to define hyperplanes within the decision nodes. We embed a multiclass strategy (one-vs-one or one-vs-rest) at the decision nodes, allowing the model to directly handle non-binary classification tasks without the need to cluster instances into two groups, as is common in other approaches from the literature. In each decision node, only the best-performing model (SVM)—the one that minimizes an impurity measure for the n-ary classification—is retained, even if the learned SVM addresses a binary classification subtask. An extensive experimental study involving 49 datasets and various state-of-the-art algorithms for oblique decision tree ensembles has been conducted. Our results show that ODTE ranks consistently above its competitors, achieving significant performance gains when hyperparameters are carefully tuned. Moreover, the oblique decision trees learned through STree are more compact than those produced by other algorithms evaluated in our experiments.
\end{abstract}




\end{frontmatter}


\section{Introduction}
\label{sec:intro}

\textit{Classification} is a problem which basically consists into assigning a label to a given instance or example from a predefined and finite set of labels. Formally, objects or instances are defined over a set of features or attributes ${\cal{X}} = \{X_1,\dots,X_n\}$, each one taking values in a predefined domain $\Omega(X_i)$ which can be of numerical or discrete nature. On the other hand, the target or \textit{class} variable $Y$ is of discrete nature and take values in a finite and mutually exclusive set of labels $\Omega(Y) = \{y_1,\dots,y_k\}$. Then, a \textit{classifier} can be viewed as a function $f:{\cal{X}} \rightarrow Y$. In machine learning (ML) or data mining (DM) \citep{DMML:Zaki} the goal is to learn this function or classifier from data. 

\textit{Supervised learning} is a main task in ML whose objective is learning a classifier from previously labelled data, i.e. objects or instances for which the correct class or label is known. This kind of data, $\mathbf{D} = \{(\vec{x}_i,y_i)\}_{i=1}^m$, usually comes from historical cases which have been previously solved, and the availability of the {\em correct} label $y_i$ for each $n$-dimensional object $\vec{x}_i$ helps the learning algorithm to \textit{supervise} the learning process, trying to minimise the error made on the given instances, whose correct label is known \citep{Classification:book}. 

A plethora of different paradigms to cope with the supervise classification problem is available in the literature: instance-based methods \citep{ibl-knn:2018}, logistic regression \citep{LogisticRegression:book}, Bayesian network classifiers \citep{BNCs:survey,BNCs:review}, decision trees \citep{DTreview:2020}, support vector machines \citep{Vapnik:book,SVM_survey:2020}, neural networks \citep{NN_classification:survey}, ensembles \citep{Ensemble:Zhou,Ensemble:Rokach}, etc. Currently, there is no doubt that from the point of view of accuracy, methods based on neural networks, especially deep learning approaches \citep{DL_AI:2021,DL:book}, and ensembles (multi-classifiers), are the most competitive ones. Nonetheless, these two outstanding approaches, while being not mutually exclusive, have their own niche of application. Thus, deep learning-based methods excel in the processing of unstructured data (image, video, etc.) and require large amounts of data to work properly. On the other hand, ensembles tend to be used more with structured (tabular) data and, in general, can work with moderate or even small amounts of data \citep{TabularData:2022}. In this paper we consider the case of  structured data, so we focus on the use of ensembles as ML paradigm.

In essence, an ensemble is a multiclassifier, that is, a set of models is learned from different samples of the data, and new cases are classified by all of them and their results aggregated to provide a final output \citep{Ensemble:survey_2018}. The use of different strategies when learning the models and aggregating their results, give rise to different ensemble methods: bagging \citep{Bagging:Breiman}, boosting \citep{Boosting:Schapire2003}, stacking \citep{Stacking:1999}, etc. These strategies are usually generic, i.e. they can be instantiated with any base classifier, although there are also ensemble models based on a specific type of classifier, such as Random Forest \citep{Breiman:RF} or Gradient Boosting \citep{GradientBoosting:2001,xgboost:2016}, which use a classification or regression trees as a base model \citep{Quinlan:C45,CART:1984}. Even in the case of generic classifiers, decision trees are also often the most widely used choice as a base classifier, due to the fact that they are models with low bias and high variance, which is exactly what is usually reduced by means of the ensemble-based aggregation, thus resulting in a more robust and stable classifier. In this study we focus on the use of decision trees as the base models for the ensemble. 

\textit{Decision Trees} (DT) \citep{DTreview:2020} are model-based classifiers whose inner structure is a tree. Basically, inner (or decision) nodes represent questions posed over the predictive attributes or features, while leaf nodes contain the class label to be returned as output. When a new instance $\vec{x}$ must be classified, the tree is traversed from the root to a leaf by choosing for each decision node the appropriate branch according to node question and the instance values for predictive attributes. Once a leaf node is reached, its associated class label is returned as outcome. The structural simplicity of DTs is one of its main advantages, as each path from the root to a leaf can be considered as a human-readable rule, which makes a DT an interpretable model able to justify its decisions \citep{XAI_Trees:2020}. Furthermore, despite the fact of its simplicity, DTs performance is competitive with other state of the art classifiers, 
having been successfully and recently applied e.g. to real-time estimation of human pose by Microsoft Kinect \citep{DT:kinect}.

\begin{figure}[htbp]
\centering
\includegraphics[width=0.7\columnwidth]{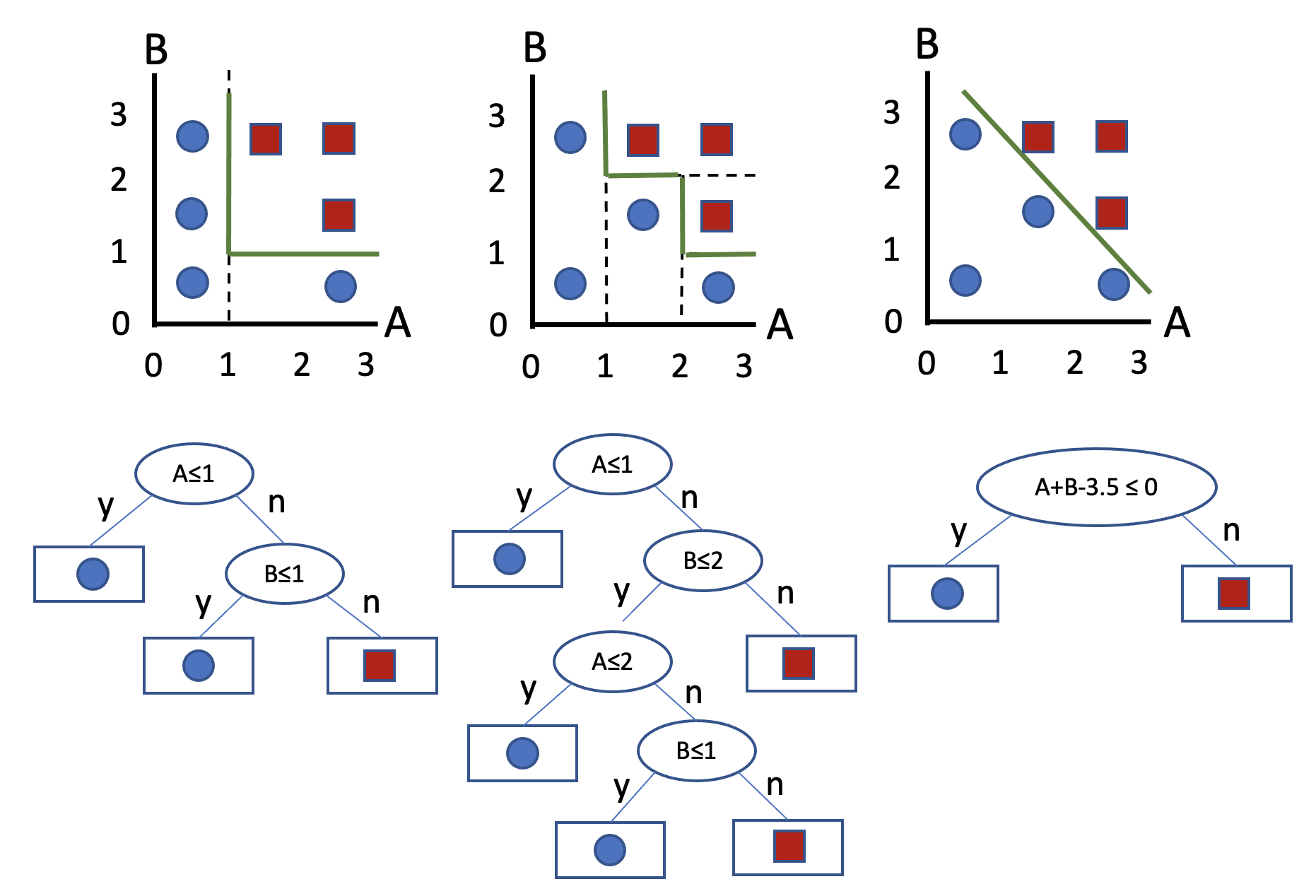}
\caption{Different decision boundaries and their corresponding decision trees: (left) axis-parallel decision boundary and simple DT; (middle) non axis-parallel decision boundary and complex DT because of the \textit{ladder} effect; (right) non axis-parallel decision boundary and simple oblique DT.}\label{fig:ladder}
\end{figure}

In literature we can find different families of decision trees, which mainly differ in the type of question/test included in the decision nodes, which in turn translate into different geometrical partitions of the representation (input) space. The most standard DT models are the so-called orthogonal ones, which consist of placing decisions of the type, $X_i \leq t$?, where $X_i$ is an attribute or feature and $t \in \Omega(X_i)$ being a threshold\footnote{In this study we focus on the case of numerical attributes, that is, $\Omega(X_i) \subseteq \mathbb{R}$. Categorical variables could be easily accommodated by some kind of standard transformation.}. 
\tcb{In a decision tree, impurity measures the heterogeneity of classes within a node. A node is pure if all its elements belong to the same class, whereas it is impure if it contains a mix of classes. The goal of a decision tree learning algorithm is to reduce impurity at each split to improve classification performance.}
Well known algorithms used to induce this type of DTs are C4.5 \citep{Quinlan:C45} and CART \citep{CART:1984}, that use information criteria like \textit{entropy} and \textit{gini} respectively to \tcb{measure the impurity of the resulting partition and} select the best splitting test ($X_i,t$) at each decision node. However, although being successful ML algorithms \citep{TopTen:Journal2007}, this type of DTs has the limitation of dealing only with decision boundaries that are parallel to the axes (features), and when this is not the case their size grow because several consecutive tests are included in order to approximate non axis-parallel boundaries (see Figure \ref{fig:ladder}).

\textit{Oblique} decision trees (ODT) \citep{OC1:journal} were introduced to capture decision boundaries not parallel to the axes. In this way it is possible to obtain more compact and generally more accurate trees. To represent these type of boundaries, the tests associated with the decision nodes are more complex, involving a function $g(\cdot)$ defined over a subset of the attributes, which on the other hand makes the resulting models lose part of their interpretability (see right part of Figure \ref{fig:ladder}). Despite this, there are a few efforts in the literature that address this issue \citep{Jungermann:2023}, and model-agnostic can also be used \citep{BONIFAZI2024:agnostic}. 
In the original proposals of ODTs the function $g(\cdot)$ was a linear combination of features, but later more complex ML models like support vector machines (SVM) or neural networks (NN) have also been used \citep{Guo:1992,Yildiz:2001,Zhang:2015}.  
Although this paper focuses on oblique decision trees, these models are part of a broader family of tree-based classifiers known as Multi-Variate Decision Trees (MVDTs). MVDTs allow combinations of features to define splits at internal nodes. A detailed analysis of these algorithms can be found in \cite{MDT_review:2021}. One notable member of the MVDT family is Functional Trees, which enable combinations of features at both inner nodes and leaf nodes. In the algorithm proposed by \cite{Gama:2004}, multivariate tests at decision nodes are created during tree growth, while functional leaves are incorporated during pruning.

Nowadays, ensemble algorithms constitute state of the art ML algorithms, usually being the first choice for practitioners to approach classification problems, especially in the case of structured (tabular) data\footnote{https://www.kdnuggets.com/2015/12/harasymiv-lessons-kaggle-machine-learning.html}. Because of the success of ensemble techniques, research on ODTs and their use as base models in ensemble classifiers, has gained attention in recent years \citep{Ganaie:ESWA2020,WODT:2019,blanco2023multiclass,Enrique_LRTO_2024}. In this article we focus on the use SVM-based ODTs as base classifiers in ensemble algorithms. In order to do so, it is necessary to deal with three different problems or choices; (1) how to deal with domains where the class variable has more than two labels?; (2) what kind of SVM algorithm to use?; and (3) what ensemble technique(s) to use for classifier design?. On the contrary to recent proposals in this research line (see Section 2) that require internal transformations to actually work with binary class problems in the decision nodes, or the use of complex and sophisticated SVM models, our proposal focuses on their ease of use by practitioners, avoiding the internal transformation of the class variable and using standard SVM algorithms that can be found in commonly used ML libraries. In fact our proposed algorithm is available as an integrated classifier in SckitLearn. The extensive experimental analysis carried out shows that the proposed ensemble, ODTE, outperforms alternative algorithms on average, standing out for its ease of use and its integration in standard software, which allows its parameters to be easily adjusted to adapt to different domains (datasets). 

In summary, our main contributions in this study are:
\begin{itemize}
\item \tcb{We introduce ODTE, an ensemble algorithm based on Oblique Decision Trees.}
\item We present STree, a novel algorithm for learning oblique decision trees that serve as the base classifiers in ODTE. This method effectively handles a multi-class target variable by producing a single model. The core idea behind the method is to embed a multiclass strategy (\tcb{One versus Rest (OvR) or One versus One (OvO)}) within each internal node, selecting only one of the learned SVMs to guide the splitting process, specifically, the one that minimizes impurity for the n-ary classification task.
%
%
\item The source code for both ODTE and STree, along with all datasets used in our study, is published on GitHub, Pypi and Zenodo to facilitte reproduciblity in future research.
\end{itemize}

In the next sections we review recent research in the design of ODT-based ensembles (Section \ref{sec:related}) and detail the main components of our proposal (Section \ref{sec:odte}). Then, in Section \ref{sec:experiments} we describe the experiments carried out to evaluate the proposed algorithm, which involve a significant benchmark of datasets and competing algorithms. Finally, our conclusions are presented in Section \ref{sec:conclu}.

\section{Related work}
\label{sec:related}

\subsection{Oblique Decision Trees} 

Constructing a Decision Tree (DT) from data typically involves a recursive partitioning process that divides the data into multiple subsets based on a selected test ({\em split}) at each internal node. This recursive partitioning stops when the data arriving at a node predominantly belongs to a single category, at which point the node becomes a leaf. Thus, the critical aspect of the DT learning process lies in determining the appropriate test or division for an internal node. In the case of axis-parallel DTs, information theory-based or statistical metrics are employed to evaluate which test most significantly diminishes the uncertainty of the class variable. Commonly utilized measures include Shannon entropy (as seen in C4.5 \citep{Quinlan:C45}) and the Gini index (as used in CART \citep{CART:1984}).

In oblique Decision Trees (DTs), more sophisticated multivariate tests are utilized, resulting in models that are generally more compact and accurate. However, selecting the appropriate test for a given internal node is more computationally demanding. In the majority of Oblique DT (ODT) algorithms, the test comprises a linear combination of the input attributes, specifically, $\beta_0 + \beta_1 X_1 + \cdots + \beta_n X_n > 0$. The objective is to identify the $\boldsymbol{\beta}$ parameters that define the hyperplane, which, in turn, produces the binary partition that most effectively reduces the uncertainty of the class variable. In the CART-LC algorithm \citep{CART:1984}, a coordinate descent method is employed for parameter optimization, whereas OC1 \citep{OC1:journal} enhances this approach by incorporating multiple restarts and random perturbations to avoid local optima. Both CART-LC and OC1 start the optimization with the optimal axis-parallel partition. WODT \citep{WODT:2019} transforms the optimization challenge by adopting a continuous and differentiable weighted information entropy function as the objective, allowing for the use of gradient descent for optimization. Furthermore, WODT also distinguishes itself by initiating with a random hyperplane. Additionally, metaheuristic algorithms have been applied to overcome local optima in this context \citep{Rafael:ODT}. 

Recent advancements in ODTs have focused on enhancing optimality and scalability, as discussed in \cite{Hu_2019}. Efforts have been made to develop sparse oblique trees that achieve low generalization error and offer rapid inference, as highlighted in \cite{Carreira_2018}. Additionally, tackling imbalanced data has become a priority, with optimization strategies now being directed towards non-linear metrics to address this issue \citep{Demirovic_2021}.

\subsection{SVM-based Oblique Decision Trees}

Beyond linear functions, advanced machine learning models such as neural networks or support vector machines (SVMs) have been explored as criteria for splits \citep{Bennett:1998,Kontschieder:2015}, enabling the application of both multivariate linear and nonlinear tests. This paper concentrates on employing SVMs \citep{SVM:1992,Vapnik:book} to develop the tests for internal nodes. The conventional SVM algorithm aims to find the optimal separating hyperplane that maximizes the margin between the training data for a binary class variable. To address non-linearly separable challenges, the input vectors might be transformed into a high-dimensional feature space, thereby converting the problem into one of linear classification. This transformation and the solution process can be efficiently combined through the use of the kernel trick, which allows for direct computation with various types of kernels (e.g., linear, polynomial).

The literature showcases a variety of Oblique Decision Tree (ODT) methodologies that incorporate Support Vector Machines (SVMs) for determining the hyperplane at each internal node. Thus, traditional SVMs equipped with linear \citep{Bennett:1998}, radial-basis function \citep{Zhang:2007}, and polynomial kernels \citep{Menkovski:2008} have been employed for this purpose. More sophisticated SVM algorithms, such as the multisurface proximal SVM (MPSVM) and the twin bounded SVM (TBSVM), are highlighted in \cite{Manwani:2012}, \cite{Zhang:2015} and \cite{Ganaie:ESWA2020} respectively. Both MPSVM and TBSVM are designed to learn two hyperplanes, each one optimized to be closer to the data samples of one class and farther from those of the opposing class; classification of instances is then achieved by evaluating the distance of instances to both hyperplanes.

The algorithms in \cite{Ganaie:ESWA2020}, \cite{Manwani:2012}, \cite{Menkovski:2008}, \cite{Zhang:2015} and \cite{blanco2023multiclass} demonstrate capability in addressing multi-class issues. In the approach discussed by \cite{Menkovski:2008}, a unique one-vs-rest binary problem is created for each class, leading to the development of $k$ SVM models. This method involves constructing a vector of length $k$ for every instance, where each dimension corresponds to the distance from the respective hyperplane. Subsequently, instances are categorized into $r$ clusters via the X-means algorithm \citep{Xmeans:2000}, with the cluster count $r$ denoting the number of branches emanating from an internal node. In the methods proposed by  \cite{Ganaie:ESWA2020} and \cite{Zhang:2015}, a method is used at internal nodes to split class labels into two groups based on Bhattacharyya distance, which then informs the binary classification tackled by MPSVM and TBSVM, respectively. \cite{Manwani:2012} present a method where the multi-class issue is simplified to a binary challenge at each internal node by opposing the class with the largest number of instances against the aggregate of other classes. In \cite{blanco2023multiclass} the authors work on multiclass categorization by introducing a mathematical programming model designed to optimize the internal structure of binary oblique decision trees. This model innovatively determines fictitious classes at the internal nodes, derived from the misclassification errors observed at the leaf nodes, thus ensuring the tree maintains a binary structure. \tcb{Recently, \cite{Chen_2019} introduced a} classification method utilizing SVM-trees combined with K-means to address the issue of imbalanced fault classification in industrial process applications, while \cite{Enrique_LRTO_2024} proposed a method to induce oblique label ranking trees to deal with the non-standard classification problem of label ranking.


\subsection{Ensembles of oblique decision trees}

Due to the rise of ensemble classifiers, oblique decision trees have naturally become prominent as base classifiers in these models. For instance, \cite{Tan:2006} introduces a new decision forest learning scheme where the base learners are Minimum Message Length (MML) oblique decision trees. \cite{zhan2024consistency} study the consistency of combining oblique decision trees with boosting and random forests, and \cite{Cantu-Paz:2003} applies evolutionary algorithms to construct ensembles of oblique decision trees.

Due to the affinity with our work, we highlight ensembles of oblique decision trees based on the use of SVMs. Notable examples include the models proposed in \cite{Ganaie:ESWA2020}, which are based on twin bounded SVMs, the models proposed in \cite{Menkovski:2008}, which incorporate heuristic techniques for enhancing the oblique decision tree construction process, and the models proposed in \cite{Zhang:2015}, which use multi-surface proximal SVMs.

Finally, the importance and rise of ensemble models based on the use of oblique decision trees is demonstrated by their penetration in various application domains. Thus, ensembles of oblique decision trees have been considered for classifying gene expression data \citep{Huynh:2018}, imbalanced fault classification in industrial processes \citep{Chen_2019}, hyperspectral data classification \citep{Poona:2016}, image classification \citep{Singh:2022}, generating explanations for neural network classifications \citep{Hada2021SparseOD}, and developing resource-efficient models for neural signal classification \citep{Zhu2020resot}, among others.


\section{Proposed method: ODTE}
\label{sec:odte}

We introduce the Oblique Decision Tree Ensemble (ODTE), a bagging-inspired ensemble with oblique decision trees as base classifiers. These trees utilize an SVM to compute hyperplanes at each inner node. In general, a binary classification problem is addressed, requiring the learned hyperplane to separate the two classes. In our description, we will treat the learned classifier as a function rather than focusing on the hyperplane inherent to the classifier (SVM). While SVM primarily deals with binary classification, a key objective of our approach is to effectively handle multi-class target variables. Unlike traditional methods that implement one-vs-rest (OvR) or one-vs-one (OvO) strategies as a separate external process, our method integrates these strategies directly at each internal node of the tree. In our proposal, we select the classifier (hyperplane) that yields the best binary partition to be included as the conditional test in the node. 

{Through the use of the OvR or OvO strategies at each decision node, we eliminate the need for clustering the set of instances into two groups, as previously required in other studies \citep{Ganaie:ESWA2020,Menkovski:2008,Zhang:2015}. Intuitively, grouping labels into a binary {\em artificial} class should result in more compact and balanced trees, however, our experiments show that segregating one label from the others as early as possible tends to produce more compact trees. Another critical aspect is our approach to select the hyperplane after facing one class label against the rest or one class label against other in a pairwise fashion. Unlike methods that typically favor the majority class label, we consider the distribution of the remaining labels across the two subsets of the partition. We then select the classifier (SVM) that results in the least impure partition, based on the distribution of the class labels in the obtained partition.

From a technical perspective, our ensemble algorithm (ODTE) (Algorithm \ref{alg:odte}), utilizes sampling with replacement to obtain the bootstrap datasets. Each (sampled) dataset is then used to train an SVM-based oblique decision tree by using our proposed STree algorithm (Algorithm \ref{alg:stree}). For inference, each ODT in the ensemble processes the given instance, and majority voting is employed to determine the final outcome (label).

In the rest of this section, we provide a detailed description of our proposed algorithm, STree\footnote{A preliminary version of STree was presented in \cite{STree:2021}}, designed to learn an SVM-based ODT from data. As usual, STree is a recursive algorithm that begins by receiving a training dataset $\mathbf{D'} = \{(\vec{x}_i,y_i)\}_{i=1}^t$. Then, the method proceeds as follows:
\begin{enumerate}
\item If the stopping condition is met, e.g., max depth has been reached or there is almost only one class label in the sample, a leaf node is created having as \textit{outcome} the more frequent class label, i.e. the mode, in the dataset $\mathbf{D'}$. 
\item Otherwise, a binary split is created using a classifier obtained by an SVM learning algorithm. Therefore, the inner classification problem(s) must be reformulated as a binary one. Basically, if the number of labels ($k'$) in the sample is greater than 2 -notice that as the tree grows in depth, not all the labels will be present- in the received set ${\mathbf{D'}}$, then a multiclass classification strategy is employed. The classifier  corresponding to the SVM whose application produces the partition with the higher information gain with respect to the class variable, is stored as the splitting test in the current node. The detailed steps of this process are outlined in algorithm \ref{alg:stree}, however, let us to highlight the main stages here:

\begin{itemize}
\item If $k^{\prime} = 2$, we are dealing with a {\em binary classification} problem. The SVM algorithm is applied to learn the classifier (hyperplane), denoted as $model$. This classifier is then used to partition the instances in $\mathbf{D'}$ into two subsets: $\mathbf{D'}^+$ and $\mathbf{D'}^-$, based on whether the output of $model(\vec{x})$ assigns the label $+$ or $-$ to the instance $\vec{x}$. Specifically, the instances above the corresponding hyperplane are assigned the label $+$, while those below the hyperplane are assigned the label $-$. The classifier ($model$) is stored in the node to serve as the decision rule during inference (line 16, Algorithm \ref{alg:stree}).
\item If $k^{\prime} > 2$ we are in a {\em multi-class} setting. In this scenario, we resort to techniques that transform a multiclass classification problem into several binary classification problems, although in the end, we will select only one of these problems (models) to be included in our tree (line 18, Algorithm \ref{alg:stree}).
\begin{itemize}

\item \textit{Learning stage}. Depending on the multiclass strategy used we have:
\begin{itemize}
\item \textit{One-vs-one} (OvO) strategy \cite[pg. 339]{Bishop:2006} defines $r=\frac{k'(k'-1)}{2}$ binary classification problems on all possible pairs of classes: $1$ (label 1 vs label 2), $2$ (label 1 vs label 3), $\dots$, $k'-1$ (label 1 vs label $k'$), $\dots$, $k'$ (label 2 vs label 3), $\dots$, $r$ (label $k'-1$ vs label $k'$). 
Let $model_j$, $j=1,\dots,r$, be the classifier learned by the SVM algorithm for the $j$-th binary classification problem (label $l$ vs label $s$; $l\neq s$), by using only the instances in $\mathbf{D'}$ labeled with $Y=l$ or $Y=s$ (line 20, Alg. \ref{alg:stree}).

\item \textit{One-vs-rest} (OvR) strategy \cite[pg. 339]{Bishop:2006}, defines $k^{\prime}$ binary classification problems by considering, respectively, each label as the positive class and the union of the remaining labels as the negative class. Let $model_j$, $j=1, \dots, k'$, be the classifier learned by the SVM algorithm for the $j$-th binary classification problem (label $j$ vs all labels minus $j$) (line 22, Alg. \ref{alg:stree}).

\end{itemize}

\item \tcb{{\em Decision stage}. Let $\mathbf{D_j'}^+$ and $\mathbf{D_j'}^-$ denote the partitions generated by one of the learned models, $model_j$, from the SVM algorithm, when applied to all instances in $\mathbf{D'}$. In the case of the OvO approach, $\mathbf{D_j'}^+$ and $\mathbf{D_j'}^-$ correspond to the instances with predicted labels $Y = l$ or $Y = s$ (represented as $+$ or $-$) for the learned model $model_j$. In the case of the OvR approach, $\mathbf{D_j'}^+$ and $\mathbf{D_j'}^-$ correspond to the instances with predicted labels $j$ or $\neg j$ (represented as $+$ or $-$) for the learned model $model_j$ (line 25, Alg. \ref{alg:stree}).} 

Let $impurity(Y,\mathbf{D})$ be a measure which evaluates the impurity of the class variable $Y$ in $\mathbf{D}$. Then, we select the classifier  $model_{b^*}$ such that 
\tcb{\[
b^* = \arg\min_{\mathbf{j}=1,..,\rho} \left (  \frac{|\mathbf{D_{j}'^{+}}|}{|\mathbf{D'}|} impurity(Y, \mathbf{D_j'^{+}}) + \frac{|\mathbf{D_j'^{-}}|}{|\mathbf{D'}|} impurity(Y, \mathbf{D_j'^{-}}) \right ),
\]}
with $\rho=k'$ or $\rho=r$ depending on the multiclass strategy, OvR or OvO, used in the learning stage. $model_{b^*}$ is then stored in the current node for future use in splitting and inference (line 25, Alg. \ref{alg:stree}).

\end{itemize}
\end{itemize}
Notice that in our pseudocode we actually maximize the gain with respect to the initial partition, which is equivalent to minimize the impurity measure. 

\item Once the classifier ($model$) and the corresponding partition ($\mathbf{D'^{+}}, \mathbf{D'^{-}}$) have been selected, two branches are created for this node: positive, for those instances being classified as $+$ by $model$ and, negative, for those instances being classified as $-$ by $model$. Two recursive calls are then launched with $\mathbf{D'^{+}}$ and $\mathbf{D'^{-}}$ as the dataset of input instances respectively (lines 29-30, Alg. \ref{alg:stree}).
\end{enumerate}

In our implementation we made the following design decisions:
\begin{itemize}
\item As there is no agreement about which kernel is better or worse for a specific domain, we allow the use of different kernels (linear, polynomial and Gaussian radial basis function). 
\item Shannon entropy is used as the impurity measure, and Information Gain is used to select the $model$ (SVM) that yields the purest partition.
\item Although a pruning stage has not yet been designed for STree, a pre-pruning strategy can be implemented by setting a maximum depth threshold for the tree.
\end{itemize}

For inference, given an instance to be classified, the tree must be traversed from the root to a leaf node, whose associated label is returned as the outcome. At each inner/decision node, the stored classifier $model$ is used, and depending on the obtained label, $+$ or $-$, the instance follows the {\em positive} or {\em negative} branch.

\algrenewcommand\algorithmicrequire{\textbf{Input:}}
\algrenewcommand\algorithmicensure{\textbf{Output:}}
\begin{algorithm}[htpb]
\caption{Oblique Decision Tree Ensemble (ODTE)}\label{alg:odte}
\begin{algorithmic}
\Require \\
$\mathbf{D} = \{(\vec{x}_i,y_i)\}_{i=1}^m$ : Data\\
$n_{trees}$ : number of trees to learn \\
\Ensure $trees$ : the ensemble 
\Function{$build\_ensemble$}{$\mathbf{D}, n_{trees}$}
    \State $trees \gets []$
    \For {$i \gets 1$ to $n_{trees}$}
        \State $\mathbf{D_i} \gets$ \Call{$bootstrap\_samples$}{$\mathbf{D}$,$m$}
        \State $trees[i] \gets$ \Call{$STree$}{$\mathbf{D_i}$}
    \EndFor
    \State \Return{trees}
\EndFunction
\end{algorithmic}
\end{algorithm}


\algrenewcommand\algorithmicrequire{\textbf{Input:}}
\algrenewcommand\algorithmicensure{\textbf{Output:}}
\begin{algorithm}[htbp]
\small
\caption{STree (Multi-class SVM-based oblique decision tree)}\label{alg:stree}
\begin{algorithmic}[1]
\Require \\
$\mathbf{D'} = \{(\vec{x}_i,y_i)\}_{i=1}^t$ : Data 
\Ensure $tree$ : the root node of an SVM-based oblique decision tree

\Function{$STree$}{$\mathbf{D'}$}
    \If {$stopping\_condition$($\{y_i\}_{i=1}^t$)}
        \State \Return {\Call {$create\_leaf\_node$}{$mode(\{y_i\}_{i=1}^t)$}} \Comment {Leaf node}
    \EndIf
    \State $k' \leftarrow num\_different\_labels(\{y_i\}_{i=1}^t)$
    \State $r \leftarrow \frac{k'(k'-1)}{2}$
    \State $Y' \leftarrow \{y'_1, \dots, y'_{k'}\}$ \Comment{set of labels}

    \State $I_{all} \leftarrow I(\{y_i\}_{i=1}^t)$ \Comment{$I(\cdot)$ is an information theory meassure}
     
    \If{(OvO)} 
        \State $n_{models} \leftarrow r$ 
    \Else \Comment{OvR \hspace*{9.5cm} $\;$}
        \State $n_{models} \leftarrow k'$ 
    \EndIf
    
    \For{$j=1$ to $n_{models}$}
        \If{($j=k'=2)$} break \Comment{Two labels, one SVM is enough}
        \EndIf
        \If{(OvO)} 
            \State Let $c_a$ and $c_b$ the pair of labels corresponding to index $j$
            \State $models[j] \gets \Call {SVM}{\mathbf{D^{\prime\downarrow{c_a}} \cup D^{\prime\downarrow{c_b}}}}$ 
        \Else \Comment{OvR \hspace*{8.5cm} $\;$}
            \State $models[j] \gets \Call {SVM}{\mathbf{D'}}$ using binary class $y'_j$ ($+$) vs rest ($-$) 
        \EndIf
        \State $\mathbf{D'^{+}}, \mathbf{D'^{-}} \leftarrow models[j](\vec{x}_i) \;\;\; \forall (\vec{x}_i) \in \mathbf{D'}$
        \State $I_j \leftarrow \frac{|\mathbf{D'^{+}}|}{|\mathbf{D'}|} I( \{y_i\}_{i=1}^{|\mathbf{D'}^+|} ) + \frac{|\mathbf{D'^{-}}|}{|\mathbf{D'}|}I(\{y_i\}_{i=1}^{|\mathbf{D'}^-|})$
        \State $IG_j \leftarrow I_{all} - I_j$
    \EndFor

        \State $b^* \leftarrow \arg \max_{j=1,\dots,n_{models}} IG_j$ 
    
    \If {$IG_{b^*} > 0$}
        \State $node \gets$ \Call {$create\_node$}{$models[b^*]$}
        \State $node.left\gets$ {\Call {$STree$}{$\mathbf{D'^{+}}$}}
        \State $node.right\gets$  {\Call {$STree$}{$\mathbf{D'^{-}}$}}
    \Else
        \State \Return {$create\_leaf\_node$}{($mode(\{y_i\}_{i=1}^t)$)} \Comment No split gain
    \EndIf
    \State \Return node
\EndFunction
\end{algorithmic}
\end{algorithm}



\section{Experimental evaluation}
\label{sec:experiments}

In this Section we describe the comprehensive experiments carried out to evaluate our proposal.


\subsection{Benchmark}

For the evaluation of our algorithm in comparison with state-of-the-art methods, we use a benchmark of 49 datasets, previously used in \cite{Ganaie:ESWA2020} to test their ensembles of oblique decision trees. Specifically, 45 of the selected datasets come from the UCI  machine learning repository \citep{UCI:ML}, while the other 4 (Id 27 to 30 in Table \ref{table:datasets}) are related to a fisheries fecundity problem (see \cite{Ganaie:ESWA2020} for the details).

The Table \ref{table:datasets} shows an identifier (Id) for each dataset, its name, the number of instances ($m$), features ($n$) and class labels ($k$).

\begin{table}[htbp]
\centering
\scriptsize
\renewcommand{\tabcolsep}{0.07cm}
\caption{Datasets used in the experiments}
\label{table:datasets}
\begin{tabular}{{rlrrrcrlrrrc}}\cline{1-5}\cline{7-11}
Id & Dataset & $m$ & $n$ & $k$ & $\;\;\;$ & Id & Dataset & $m$ & $n$ & $k$\\ \cline{1-5}\cline{7-11}
1 & balance-scale & 625 & 4 & 3 & & 26 & musk-1 & 476 & 166 & 2 \\ 
2 & balloons & 16 & 4 & 2 & & 27 & oocytes\_merluccius\_nucleus\_4d & 1022 & 41 & 2 \\ 
3 & breast-cancer & 286 & 9 & 2 & & 28 & oocytes\_merluccius\_states\_2f & 1022 & 25 & 3 \\ 
4 & breast-cancer-wisc & 699 & 9 & 2 & & 29 & oocytes\_trisopterus\_nucleus\_2f & 912 & 25 & 2  \\ 
5 & breast-cancer-wisc-diag & 569 & 30 & 2 & & 30 & oocytes\_trisopterus\_states\_5b & 912 & 32 & 3  \\ 
6 & breast-cancer-wisc-prog & 198 & 33 & 2 & & 31 & parkinsons & 195 & 22 & 2 \\ 
7 & cardiotocography-10clases & 2126 & 21 & 10 & & 32 & pima & 768 & 8 & 2 \\ 
8 & cardiotocography-3clases & 2126 & 21 & 3 & & 33 & pittsburg-bridges-MATERIAL & 106 & 7 & 3 \\ 
9 & conn-bench-sonar-mines-rocks & 208 & 60 & 2 & & 34 & pittsburg-bridges-REL-L & 103 & 7 & 3\\ 
10 & cylinder-bands & 512 & 35 & 2 & & 35 & pittsburg-bridges-SPAN & 92 & 7 & 3 \\ 
11 & dermatology & 366 & 34 & 6 & & 36 & pittsburg-bridges-T-OR-D & 102 & 7 & 2 \\ 
12 & echocardiogram & 131 & 10 & 2 & & 37 & planning & 182 & 12 & 2 \\ 
13 & fertility & 100 & 9 & 2 & & 38 & post-operative & 90 & 8 & 3\\ 
14 & haberman-survival & 306 & 3 & 2 & & 39 & seeds & 210 & 7 & 3\\ 
15 & heart-hungarian & 294 & 12 & 2 & & 40 & statlog-australian-credit & 690 & 14 & 2 \\ 
16 & hepatitis & 155 & 19 & 2 & & 41 & statlog-german-credit & 1000 & 24 & 2\\ 
17 & ilpd-indian-liver & 583 & 9 & 2 & & 42 & statlog-heart & 270 & 13 & 2\\ 
18 & ionosphere & 351 & 33 & 2 & & 43 & statlog-image & 2310 & 18 & 7\\ 
19 & iris & 150 & 4 & 3 & & 44 & statlog-vehicle & 846 & 18 & 4  \\ 
20 & led-display & 1000 & 7 & 10 & & 45 & synthetic-control & 600 & 60 & 6 \\ 
21 & libras & 360 & 90 & 15 & &  46 & tic-tac-toe & 958 & 9 & 2 \\ 
22 & low-res-spect & 531 & 100 & 9 & & 47 & vertebral-column-2clases & 310 & 6 & 2 \\
23 & lymphography & 148 & 18 & 4 & & 48 & wine & 178 & 13 & 3 \\
24 & mammographic & 961 & 5 & 2 & & 49 & zoo & 101 & 16 & 7\\ 
25 & molec-biol-promoter & 106 & 57 & 2 & & \\ \cline{1-5}\cline{7-11}
\end{tabular}
\end{table}

\begin{table}[ht]
    \centering
    \begin{tabular}{|c|c|}
        \hline
        Hyperparameter & Value \\
        \hline
        C & 1 \\
        kernel & linear \\
        multiclass\_strategy & OvO \\
        splitter & random \\
        max\_features & None \\
        max\_iter & 1e5 \\
        \hline
    \end{tabular}
    \caption{Default STree hyperparameters.}
    \label{tab:stree_default}
\end{table}


\subsection{Algorithms}

The following algorithms have been included in the comparison:

\begin{itemize}
    
    \item {\sf ODTE}. This is our proposed algorithm with default hyperparameters: 100 trees, STree as base estimator, don't limit number of features and don't limit the size of the bootstrap. Also default hyperparameters are used for STree.

    \item {\sf ODTE\textsuperscript{T}}. This is the tuned version of our proposed algorithm. In this case the hyperparameters for the base estimators were optimized for each dataset using a gridsearch. The obtained values for each hyperparameter and dataset are available at {\sf https://t.ly/rEZLH}.
    
    \item {\sf TBRaF} \citep{Ganaie:ESWA2020}. Twin Bounded Random Forest uses the fundamentals of Random Forests such as Bagging and random subspaces. Each tree is built with a bootstrap of the training data with a random subspace of features, and uses Twin Bonded Support Vector Machine \citep{Shao_2011} to generate the hyperplane needed, in this case 2 hyperplanes are generated, for splitting at each non-leaf node. The algorithm is run with a default parameter setting $c1=0.25$ and $c3=0.0625$.
    
    \item {\sf TBRaF\textsuperscript{T}}. This is the tuned version of the previous model. The hyperparameters selected by the authors in the provided code are selected for each dataset (see {\sf https://t.ly/zi-Ma}).
    
    \item {\sf TBRoF} \citep{Ganaie:ESWA2020}. Twin Bounded Rotated Forest uses Principal Component Analysis (PCA) to rotate the dataset \citep{Rodriguez_2006} and different rotation matrices are used to build each tree. The whole set of features are used to compute the hyperplanes in each node. The algorithm is run with a default parameter setting $c1=0.25$ and $c3=0.0625$.

    \item {\sf TBRoF\textsuperscript{T}}. This is the tuned version of the previous model. The hyperparameters selected by the authors in the provided code are selected for each dataset (see {\sf https://t.ly/WI8sS}).
    
    \item {\sf TBRRoF} \citep{Ganaie:ESWA2020}. Twin Bounded subrotation Forest Algorithm mixes PCA rotation and random subspaces, this way the rotation is applied only to the subset of features that is assigned for each tree. The algorithm is run with a default parameter setting $c1=0.25$ and $c3=0.0625$.

    \item {\sf TBRRoF\textsuperscript{T}}. This is the tuned version of the previous model. The hyperparameters selected by the authors in the provided code are selected for each dataset (see {\sf  https://t.ly/gfr6T}).
    
    \item {\sf BaggingJ48-SVMODT}. We have used Weka \citep{Weka:tool} Bagging Classifier to wrap J48-SVMODT \citep{Menkovski:2008} as base classifier. The {\sf Hedge} algorithm proposed in  \cite{Menkovski:2008} that also uses J48-SVMODT as base classifier, has not been included in the experiments since it is clearly surpassed by Bag\-gingJ48-SVMODT. 

    \item{\sf PBC4cip}. This is an ensemble based on multivariate decision trees that uses contrast patterns and a filtering algorithm \citep{PBC4cip:2017}. This algorithm\footnote{The implementation provided by PBC4cip algorithm has been used: {\sf https://sites.google.com/view/leocanetesifuentes/home}} has been compared against other multivariate decision trees in \cite{pbc4cip:2019}, demonstrating very good performance, particularly with imbalanced datasets.
    
    \item {\sf BaggingWODT}. We have used bagging classifier from scikit-learn \citep{scikit-learn} to wrap WODT \citep{WODT:2019} as base classifier.

    \item {\sf RandomForest} \citep{Breiman:RF} and {\sf XGBoost} \citep{xgboost:2016}. State-of-the-art ensemble algorithms taken from scikit-learn.

 
\end{itemize}

As a common parameter, 100 trees are learned as base classifiers in all the ensemble algorithms, except for {\sf PBC4cip} that uses 150 trees by default.



\subsection{Reproducibility}\label{ss:reproducibility}

ODTE and STree have been implemented in {\tt python} as {\tt scikit-learn} classifiers. For the sake of reproducibility and boosting future comparisons, we provide the code of both classifiers in: \href{https://github.com/Doctorado-ML/Odte}{ODTE}\footnote{{\sf https://github.com/Doctorado-ML/Odte}} and \href{https://github.com/doctorado-ml/stree}{STree}\footnote{{\sf https://github.com/doctorado-ml/stree}}. Furthermore, to make easier reproducing our experiments, the datasets have been uploaded to \href{https://zenodo.org/records/13119174}{Zenodo}\footnote{{\sf https://zenodo.org/records/13119174}}.

For the rest of algorithms we have used publicly available versions of RandomForest ({\tt python/scikit-learn}) and XGBoost pre-built binary wheels for Python from PyPI (Python Package Index) have been used. The code for WODT ({\tt python}), J48SVM-ODT ({\tt Java/Weka}), PBC4cip ({\sf Java/Weka}) and TBSVM-ODT ({\sf Matlab}) have been provided to us by their authors. To create the ensembles for the base algorithms WODT and J48SVM-ODT we have wrappered them by using the public implementation of {\em bagging} available in \href{https://scikit-learn.org/stable/modules/generated/sklearn.ensemble.BaggingClassifier.html}{scikit-learn} and \href{https://weka.sourceforge.io/doc.stable-3-8/weka/classifiers/meta/Bagging.html}{Weka} respectively. 

All the experiments have been run in an Intel(R) Xeon(R) CPU E5645 at 2.40GHz running Linux operating system.


\subsection{Results and analysis} \label{sec:results}

To evaluate the performance of each pair (algorithm, dataset), we conducted a ten-times five-fold cross validation (10$\times$5cv). The same 10 seeds were used across all pairs to randomize the data prior to cross-validation. Since no severe imbalance is present in any dataset (see \cite[Table A1]{Ganaie:ESWA2020}), accuracy was used to compare the performance of the tested algorithms. 

We distinguish two scenarios: all the algorithms are run by using a default setting for the hyperparameters (see Table \ref{tab:stree_default}); and some of the algorithms are run by using hyperparameter tuning  In this second case only our proposed algorithm ({\sf ODTE\textsuperscript{T}}) and the algorithms for which the hyperparameter tuning is available ({\sc TBRaF\textsuperscript{T}}, {\sf TBRoF\textsuperscript{T}} and {\sf TBRRoF\textsuperscript{T}}) are considered. The mean and standard deviation over the 50 runs of the 10$\times$5cv are reported, respectively, in Tables \ref{tab:results_accuracy} \& \ref{tab:results_accuracy@}.


\begin{table}[htbp] 
\centering 
\tiny 
\renewcommand{\arraystretch }{1.2} 
\renewcommand{\tabcolsep }{0.07cm} 
\caption{Accuracy results(mean $\pm$ std) for all the algorithms and datasets} 
\label{tab:results_accuracy}
\begin{tabular} {{rccccccccc}}
\hline 

& Bagg-SVMODT& BaggingWodt& Odte& PBC4cip& RandomForest& TBRRoF& TBRaF& TBRoF& XGBoost\\
\hline
1 & 0.9498$\pm$0.018& \bfseries0.9566$\pm$0.018& 0.9238$\pm$0.022& 0.9067$\pm$0.025& 0.8376$\pm$0.020& 0.9028$\pm$0.021& 0.9023$\pm$0.024& 0.6981$\pm$0.231& 0.8632$\pm$0.026\\
2 & 0.5983$\pm$0.219& 0.6700$\pm$0.247& 0.6450$\pm$0.304& \bfseries0.6900$\pm$0.271& 0.6233$\pm$0.285& 0.6388$\pm$0.236& 0.5938$\pm$0.230& 0.5837$\pm$0.241& 0.6300$\pm$0.305\\
3 & 0.7227$\pm$0.051& 0.7026$\pm$0.053& \bfseries0.7391$\pm$0.026& 0.6195$\pm$0.072& 0.7279$\pm$0.046& 0.7358$\pm$0.040& 0.7378$\pm$0.046& 0.6895$\pm$0.050& 0.6689$\pm$0.054\\
4 & 0.9685$\pm$0.012& 0.9631$\pm$0.012& 0.9671$\pm$0.012& \bfseries0.9704$\pm$0.013& 0.9680$\pm$0.012& 0.9679$\pm$0.014& 0.9682$\pm$0.012& 0.9404$\pm$0.016& 0.9575$\pm$0.014\\
5 & 0.9652$\pm$0.018& \bfseries0.9768$\pm$0.012& 0.9717$\pm$0.015& 0.9633$\pm$0.014& 0.9597$\pm$0.018& 0.9724$\pm$0.014& 0.9718$\pm$0.013& 0.9575$\pm$0.016& 0.9645$\pm$0.019\\
6 & \bfseries0.8039$\pm$0.061& 0.7991$\pm$0.051& 0.7947$\pm$0.049& 0.6115$\pm$0.066& 0.7936$\pm$0.030& 0.8007$\pm$0.058& 0.7880$\pm$0.061& 0.7515$\pm$0.060& 0.7929$\pm$0.045\\
7 & 0.8692$\pm$0.014& 0.8344$\pm$0.020& 0.8378$\pm$0.017& 0.8686$\pm$0.016& 0.8714$\pm$0.018& 0.6963$\pm$0.025& 0.6568$\pm$0.058& 0.7837$\pm$0.018& \bfseries0.8841$\pm$0.015\\
8 & 0.9442$\pm$0.011& 0.9235$\pm$0.013& 0.9149$\pm$0.013& 0.9238$\pm$0.012& 0.9428$\pm$0.011& 0.8594$\pm$0.019& 0.8737$\pm$0.018& 0.8930$\pm$0.014& \bfseries0.9518$\pm$0.008\\
9 & 0.7952$\pm$0.071& \bfseries0.8469$\pm$0.057& 0.7779$\pm$0.060& 0.8352$\pm$0.056& 0.8117$\pm$0.053& 0.8378$\pm$0.047& 0.8213$\pm$0.051& 0.7517$\pm$0.053& 0.8413$\pm$0.055\\
10 & 0.7895$\pm$0.036& 0.7605$\pm$0.045& 0.7441$\pm$0.043& 0.7812$\pm$0.043& 0.7941$\pm$0.039& 0.7591$\pm$0.039& 0.7338$\pm$0.045& 0.6790$\pm$0.043& \bfseries0.7992$\pm$0.031\\
11 & 0.9667$\pm$0.025& 0.9686$\pm$0.018& 0.9702$\pm$0.016& 0.9708$\pm$0.017& 0.9751$\pm$0.014& \bfseries0.9755$\pm$0.015& 0.9742$\pm$0.015& 0.9664$\pm$0.018& 0.9659$\pm$0.019\\
12 & 0.8374$\pm$0.065& 0.7722$\pm$0.074& 0.8306$\pm$0.052& 0.7800$\pm$0.070& 0.8208$\pm$0.062& \bfseries0.8425$\pm$0.056& 0.8415$\pm$0.055& 0.7448$\pm$0.070& 0.8033$\pm$0.074\\
13 & 0.8720$\pm$0.069& 0.8570$\pm$0.056& 0.8644$\pm$0.035& 0.6690$\pm$0.089& 0.8710$\pm$0.038& \bfseries0.8800$\pm$0.060& \bfseries0.8800$\pm$0.059& 0.7890$\pm$0.088& 0.8430$\pm$0.052\\
14 & 0.7244$\pm$0.053& 0.6807$\pm$0.052& 0.7301$\pm$0.016& 0.6434$\pm$0.050& 0.6968$\pm$0.049& 0.7291$\pm$0.042& \bfseries0.7305$\pm$0.043& 0.7252$\pm$0.046& 0.6599$\pm$0.048\\
15 & 0.8000$\pm$0.052& 0.8112$\pm$0.049& 0.8157$\pm$0.045& 0.8149$\pm$0.051& 0.8130$\pm$0.039& 0.8138$\pm$0.038& \bfseries0.8179$\pm$0.041& 0.7598$\pm$0.045& 0.7939$\pm$0.048\\
16 & 0.7948$\pm$0.071& 0.8239$\pm$0.086& 0.8171$\pm$0.065& 0.7729$\pm$0.075& \bfseries0.8311$\pm$0.054& 0.8139$\pm$0.050& 0.8134$\pm$0.057& 0.7618$\pm$0.062& 0.7858$\pm$0.058\\
17 & 0.6916$\pm$0.037& 0.7091$\pm$0.031& 0.7136$\pm$0.008& 0.6764$\pm$0.041& 0.7032$\pm$0.038& \bfseries0.7137$\pm$0.034& 0.7078$\pm$0.031& 0.6771$\pm$0.038& 0.6938$\pm$0.039\\
18 & 0.9270$\pm$0.029& 0.9120$\pm$0.034& 0.9054$\pm$0.030& \bfseries0.9379$\pm$0.026& 0.9325$\pm$0.024& 0.9178$\pm$0.025& 0.9320$\pm$0.024& 0.8642$\pm$0.033& 0.9231$\pm$0.028\\
19 & 0.9433$\pm$0.037& 0.9487$\pm$0.031& 0.9593$\pm$0.041& 0.9653$\pm$0.033& 0.9473$\pm$0.040& 0.9656$\pm$0.027& \bfseries0.9665$\pm$0.025& 0.9584$\pm$0.027& 0.9413$\pm$0.037\\
20 & 0.7209$\pm$0.028& 0.7071$\pm$0.030& 0.7146$\pm$0.027& \bfseries0.7287$\pm$0.024& 0.7114$\pm$0.026& 0.3765$\pm$0.090& 0.7044$\pm$0.030& 0.6682$\pm$0.102& 0.7150$\pm$0.026\\
21 & 0.7586$\pm$0.051& 0.8314$\pm$0.041& 0.8729$\pm$0.041& 0.8383$\pm$0.045& 0.8206$\pm$0.039& \bfseries0.8839$\pm$0.035& 0.8578$\pm$0.040& 0.7007$\pm$0.052& 0.7547$\pm$0.055\\
22 & 0.8787$\pm$0.033& 0.8874$\pm$0.030& \bfseries0.9111$\pm$0.023& 0.8913$\pm$0.027& 0.8952$\pm$0.025& 0.8993$\pm$0.025& 0.8975$\pm$0.024& 0.8161$\pm$0.030& 0.8891$\pm$0.022\\
23 & 0.8062$\pm$0.076& \bfseries0.8520$\pm$0.058& 0.8334$\pm$0.066& 0.8410$\pm$0.054& 0.8451$\pm$0.053& 0.8332$\pm$0.059& 0.8331$\pm$0.063& 0.7612$\pm$0.071& 0.8462$\pm$0.058\\
24 & 0.8279$\pm$0.023& 0.7827$\pm$0.023& \bfseries0.8311$\pm$0.026& 0.8185$\pm$0.021& 0.7997$\pm$0.030& 0.8210$\pm$0.022& 0.8291$\pm$0.022& 0.8120$\pm$0.022& 0.7983$\pm$0.027\\
25 & 0.8374$\pm$0.079& 0.8180$\pm$0.083& 0.7864$\pm$0.081& 0.8786$\pm$0.076& \bfseries0.8872$\pm$0.067& 0.8194$\pm$0.075& 0.8137$\pm$0.073& 0.6292$\pm$0.083& 0.8605$\pm$0.073\\
26 & 0.8948$\pm$0.029& \bfseries0.8958$\pm$0.036& 0.8611$\pm$0.039& 0.8922$\pm$0.032& 0.8939$\pm$0.032& 0.8937$\pm$0.026& 0.8780$\pm$0.032& 0.8129$\pm$0.036& 0.8897$\pm$0.031\\
27 & 0.8047$\pm$0.024& 0.8024$\pm$0.021& 0.8079$\pm$0.022& \bfseries0.8327$\pm$0.023& 0.8047$\pm$0.021& 0.8251$\pm$0.023& 0.8179$\pm$0.022& 0.8174$\pm$0.022& 0.8115$\pm$0.025\\
28 & 0.9172$\pm$0.021& 0.9189$\pm$0.017& 0.9242$\pm$0.013& 0.9045$\pm$0.016& 0.9242$\pm$0.018& 0.9237$\pm$0.015& 0.9226$\pm$0.015& 0.9165$\pm$0.017& \bfseries0.9281$\pm$0.018\\
29 & \bfseries0.8317$\pm$0.027& 0.8086$\pm$0.024& 0.8087$\pm$0.024& 0.8133$\pm$0.029& 0.8102$\pm$0.023& 0.8302$\pm$0.024& 0.8271$\pm$0.025& 0.7843$\pm$0.030& 0.8248$\pm$0.028\\
30 & 0.9246$\pm$0.018& 0.9236$\pm$0.017& 0.9274$\pm$0.017& 0.9211$\pm$0.018& 0.9145$\pm$0.020& 0.9046$\pm$0.019& 0.9057$\pm$0.019& 0.9277$\pm$0.015& \bfseries0.9282$\pm$0.017\\
31 & 0.9021$\pm$0.040& 0.9277$\pm$0.035& 0.9031$\pm$0.050& 0.8600$\pm$0.065& 0.9047$\pm$0.051& \bfseries0.9297$\pm$0.036& 0.9141$\pm$0.036& 0.8735$\pm$0.045& 0.9138$\pm$0.043\\
32 & 0.7624$\pm$0.032& 0.7459$\pm$0.031& \bfseries0.7725$\pm$0.025& 0.7462$\pm$0.032& 0.7626$\pm$0.023& 0.7655$\pm$0.028& 0.7646$\pm$0.029& 0.7358$\pm$0.030& 0.7332$\pm$0.030\\
33 & 0.8530$\pm$0.067& 0.8101$\pm$0.074& 0.8372$\pm$0.050& 0.7974$\pm$0.073& 0.8373$\pm$0.052& 0.8527$\pm$0.053& \bfseries0.8651$\pm$0.058& 0.7952$\pm$0.079& 0.8320$\pm$0.058\\
34 & 0.6740$\pm$0.101& 0.6690$\pm$0.088& 0.6851$\pm$0.081& 0.6721$\pm$0.115& \bfseries0.6945$\pm$0.091& 0.6779$\pm$0.084& 0.6899$\pm$0.078& 0.5911$\pm$0.097& 0.6592$\pm$0.101\\
35 & 0.6540$\pm$0.113& 0.6481$\pm$0.112& \bfseries0.6854$\pm$0.089& 0.6247$\pm$0.105& 0.6420$\pm$0.080& 0.6570$\pm$0.089& 0.6770$\pm$0.079& 0.5811$\pm$0.091& 0.6673$\pm$0.091\\
36 & 0.8474$\pm$0.076& 0.8579$\pm$0.078& 0.8638$\pm$0.051& 0.7877$\pm$0.087& 0.8501$\pm$0.045& 0.8628$\pm$0.056& \bfseries0.8671$\pm$0.064& 0.8264$\pm$0.075& 0.8502$\pm$0.050\\
37 & 0.7037$\pm$0.056& 0.6696$\pm$0.072& 0.7144$\pm$0.010& 0.4686$\pm$0.070& 0.6848$\pm$0.043& \bfseries0.7177$\pm$0.057& 0.7139$\pm$0.060& 0.6137$\pm$0.074& 0.6169$\pm$0.070\\
38 & 0.6767$\pm$0.081& 0.5944$\pm$0.087& 0.7036$\pm$0.054& 0.3989$\pm$0.100& 0.6278$\pm$0.094& 0.6920$\pm$0.085& \bfseries0.7072$\pm$0.084& 0.5345$\pm$0.101& 0.5689$\pm$0.085\\
39 & 0.9281$\pm$0.044& 0.9400$\pm$0.041& 0.9357$\pm$0.028& 0.9376$\pm$0.031& 0.9252$\pm$0.034& \bfseries0.9478$\pm$0.029& 0.9424$\pm$0.030& 0.9417$\pm$0.031& 0.9348$\pm$0.038\\
40 & 0.6765$\pm$0.035& 0.6170$\pm$0.034& \bfseries0.6783$\pm$0.003& 0.5549$\pm$0.036& 0.6702$\pm$0.024& 0.6454$\pm$0.030& 0.6688$\pm$0.030& 0.5800$\pm$0.040& 0.6336$\pm$0.032\\
41 & \bfseries0.7669$\pm$0.022& 0.7597$\pm$0.027& 0.7605$\pm$0.018& 0.7245$\pm$0.027& 0.7642$\pm$0.021& 0.7468$\pm$0.026& 0.7407$\pm$0.030& 0.7263$\pm$0.025& 0.7602$\pm$0.023\\
42 & 0.8359$\pm$0.043& 0.8267$\pm$0.045& \bfseries0.8389$\pm$0.043& 0.8285$\pm$0.042& 0.8348$\pm$0.050& 0.8333$\pm$0.042& 0.8369$\pm$0.039& 0.7514$\pm$0.054& 0.8085$\pm$0.047\\
43 & 0.9774$\pm$0.006& 0.9713$\pm$0.007& 0.9639$\pm$0.008& 0.9756$\pm$0.007& 0.9784$\pm$0.008& 0.9340$\pm$0.024& 0.9420$\pm$0.021& 0.9425$\pm$0.070& \bfseries0.9819$\pm$0.007\\
44 & 0.7596$\pm$0.028& 0.7625$\pm$0.028& \bfseries0.8129$\pm$0.030& 0.7786$\pm$0.026& 0.7513$\pm$0.026& 0.7360$\pm$0.038& 0.7547$\pm$0.030& 0.7901$\pm$0.027& 0.7689$\pm$0.027\\
45 & 0.9613$\pm$0.022& 0.9850$\pm$0.010& 0.9865$\pm$0.009& 0.9862$\pm$0.010& 0.9868$\pm$0.009& \bfseries0.9922$\pm$0.007& 0.9903$\pm$0.008& 0.9518$\pm$0.019& 0.9753$\pm$0.012\\
46 & 0.9833$\pm$0.008& 0.9855$\pm$0.009& 0.9833$\pm$0.009& 0.9867$\pm$0.010& \bfseries0.9890$\pm$0.009& 0.8797$\pm$0.029& 0.8656$\pm$0.025& 0.9833$\pm$0.008& 0.9888$\pm$0.008\\
47 & 0.8548$\pm$0.043& 0.8361$\pm$0.043& \bfseries0.8565$\pm$0.039& 0.8223$\pm$0.038& 0.8297$\pm$0.036& 0.8434$\pm$0.038& 0.8426$\pm$0.037& 0.8190$\pm$0.041& 0.8271$\pm$0.036\\
48 & 0.9792$\pm$0.020& 0.9786$\pm$0.022& 0.9747$\pm$0.021& 0.9703$\pm$0.033& 0.9797$\pm$0.024& \bfseries0.9846$\pm$0.017& 0.9818$\pm$0.020& 0.9797$\pm$0.026& 0.9635$\pm$0.031\\
49 & 0.9317$\pm$0.047& 0.9575$\pm$0.038& 0.9579$\pm$0.048& \bfseries0.9743$\pm$0.035& 0.9607$\pm$0.040& 0.9468$\pm$0.045& 0.9346$\pm$0.056& 0.9307$\pm$0.052& 0.9585$\pm$0.035\\
\hline
Average & 0.8346$\pm$0.100& 0.8303$\pm$0.108& \bfseries0.8391$\pm$0.095& 0.8093$\pm$0.138& 0.8347$\pm$0.102& 0.8261$\pm$0.116& 0.8306$\pm$0.099& 0.7871$\pm$0.121& 0.8256$\pm$0.111\\
\hline 
\end{tabular}
\end{table}


\begin{table}[p] 
\centering 
\tiny 
\renewcommand{\arraystretch }{1.2} 
\renewcommand{\tabcolsep }{0.07cm} 
\caption{Accuracy results(mean $\pm$ std) for all the algorithms and datasets with optimized hyperparameters} 
\label{tab:results_accuracy@}
\begin{tabular} {{rcccc}}
\hline 

& ODTE\textsuperscript{T}& TBRRoF\textsuperscript{T}& TBRaF\textsuperscript{T}& TBRoF\textsuperscript{T}\\
\hline
1 & \bfseries0.9750$\pm$0.013& 0.9044$\pm$0.021& 0.9023$\pm$0.024& 0.7787$\pm$0.188\\
2 & \bfseries0.7250$\pm$0.286& 0.6288$\pm$0.238& 0.6000$\pm$0.232& 0.6188$\pm$0.243\\
3 & \bfseries0.7517$\pm$0.033& 0.7347$\pm$0.053& 0.7331$\pm$0.045& 0.7227$\pm$0.043\\
4 & \bfseries0.9707$\pm$0.011& 0.9688$\pm$0.012& 0.9690$\pm$0.012& 0.9618$\pm$0.015\\
5 & \bfseries0.9761$\pm$0.012& 0.9741$\pm$0.013& 0.9714$\pm$0.013& 0.9734$\pm$0.013\\
6 & 0.8031$\pm$0.042& \bfseries0.8115$\pm$0.053& 0.7999$\pm$0.054& 0.8043$\pm$0.057\\
7 & \bfseries0.8405$\pm$0.017& 0.8327$\pm$0.017& 0.8324$\pm$0.016& 0.7827$\pm$0.019\\
8 & \bfseries0.9264$\pm$0.012& 0.9222$\pm$0.012& 0.9197$\pm$0.011& 0.8954$\pm$0.016\\
9 & \bfseries0.8587$\pm$0.048& 0.8357$\pm$0.049& 0.8216$\pm$0.054& 0.7905$\pm$0.054\\
10 & \bfseries0.7842$\pm$0.041& 0.7728$\pm$0.036& 0.7510$\pm$0.038& 0.7148$\pm$0.038\\
11 & \bfseries0.9806$\pm$0.014& 0.9752$\pm$0.015& 0.9741$\pm$0.016& 0.9732$\pm$0.015\\
12 & \bfseries0.8550$\pm$0.050& 0.8507$\pm$0.053& 0.8496$\pm$0.052& 0.8292$\pm$0.056\\
13 & 0.8806$\pm$0.021& \bfseries0.8810$\pm$0.065& 0.8804$\pm$0.063& 0.8800$\pm$0.060\\
14 & 0.7350$\pm$0.022& \bfseries0.7374$\pm$0.041& 0.7356$\pm$0.045& 0.7326$\pm$0.044\\
15 & \bfseries0.8232$\pm$0.039& 0.8183$\pm$0.037& 0.8203$\pm$0.041& 0.8068$\pm$0.039\\
16 & \bfseries0.8425$\pm$0.052& 0.8180$\pm$0.054& 0.8195$\pm$0.055& 0.8099$\pm$0.056\\
17 & \bfseries0.7316$\pm$0.017& 0.7133$\pm$0.033& 0.7143$\pm$0.033& 0.7072$\pm$0.034\\
18 & \bfseries0.9498$\pm$0.023& 0.9471$\pm$0.021& 0.9321$\pm$0.025& 0.9069$\pm$0.030\\
19 & 0.9653$\pm$0.036& 0.9653$\pm$0.025& 0.9630$\pm$0.025& \bfseries0.9772$\pm$0.020\\
20 & \bfseries0.7275$\pm$0.030& 0.6862$\pm$0.053& 0.7126$\pm$0.023& 0.7029$\pm$0.026\\
21 & 0.8727$\pm$0.044& \bfseries0.8851$\pm$0.034& 0.8527$\pm$0.037& 0.7556$\pm$0.047\\
22 & \bfseries0.9115$\pm$0.021& 0.8994$\pm$0.024& 0.9007$\pm$0.025& 0.8686$\pm$0.030\\
23 & \bfseries0.8636$\pm$0.058& 0.8559$\pm$0.058& 0.8485$\pm$0.058& 0.7734$\pm$0.066\\
24 & \bfseries0.8353$\pm$0.028& 0.8261$\pm$0.023& 0.8296$\pm$0.021& 0.8130$\pm$0.020\\
25 & \bfseries0.9014$\pm$0.067& 0.8146$\pm$0.064& 0.8230$\pm$0.071& 0.7480$\pm$0.079\\
26 & \bfseries0.9332$\pm$0.031& 0.8939$\pm$0.028& 0.8807$\pm$0.030& 0.8629$\pm$0.030\\
27 & \bfseries0.8411$\pm$0.026& 0.8389$\pm$0.020& 0.8391$\pm$0.020& 0.8256$\pm$0.023\\
28 & \bfseries0.9326$\pm$0.015& 0.9270$\pm$0.015& 0.9238$\pm$0.016& 0.9195$\pm$0.016\\
29 & \bfseries0.8557$\pm$0.021& 0.8365$\pm$0.024& 0.8345$\pm$0.023& 0.8049$\pm$0.025\\
30 & 0.9309$\pm$0.019& \bfseries0.9323$\pm$0.016& 0.9279$\pm$0.017& 0.9304$\pm$0.016\\
31 & \bfseries0.9344$\pm$0.032& 0.9310$\pm$0.036& 0.9143$\pm$0.041& 0.8948$\pm$0.046\\
32 & \bfseries0.7727$\pm$0.024& 0.7685$\pm$0.030& 0.7633$\pm$0.028& 0.7675$\pm$0.028\\
33 & \bfseries0.8732$\pm$0.052& 0.8602$\pm$0.054& 0.8667$\pm$0.057& 0.8014$\pm$0.077\\
34 & \bfseries0.7061$\pm$0.090& 0.6904$\pm$0.076& 0.6956$\pm$0.083& 0.6185$\pm$0.094\\
35 & \bfseries0.6942$\pm$0.083& 0.6924$\pm$0.085& 0.6865$\pm$0.085& 0.6222$\pm$0.095\\
36 & 0.8686$\pm$0.042& 0.8702$\pm$0.056& 0.8650$\pm$0.058& \bfseries0.8714$\pm$0.063\\
37 & \bfseries0.7309$\pm$0.018& 0.7202$\pm$0.058& 0.7146$\pm$0.054& 0.7131$\pm$0.059\\
38 & \bfseries0.7121$\pm$0.028& 0.7109$\pm$0.084& 0.7074$\pm$0.087& 0.7115$\pm$0.085\\
39 & \bfseries0.9605$\pm$0.025& 0.9468$\pm$0.027& 0.9439$\pm$0.030& 0.9507$\pm$0.042\\
40 & 0.6783$\pm$0.003& \bfseries0.6783$\pm$0.031& 0.6783$\pm$0.029& 0.6783$\pm$0.031\\
41 & \bfseries0.7682$\pm$0.020& 0.7578$\pm$0.027& 0.7548$\pm$0.025& 0.7561$\pm$0.025\\
42 & \bfseries0.8481$\pm$0.041& 0.8367$\pm$0.039& 0.8374$\pm$0.040& 0.8250$\pm$0.043\\
43 & 0.9711$\pm$0.007& \bfseries0.9760$\pm$0.007& 0.9698$\pm$0.008& 0.9568$\pm$0.033\\
44 & \bfseries0.8133$\pm$0.026& 0.7863$\pm$0.025& 0.7765$\pm$0.027& 0.8033$\pm$0.044\\
45 & 0.9910$\pm$0.008& \bfseries0.9920$\pm$0.007& 0.9913$\pm$0.007& 0.9807$\pm$0.014\\
46 & \bfseries0.9939$\pm$0.008& 0.9857$\pm$0.007& 0.9799$\pm$0.012& 0.9833$\pm$0.008\\
47 & \bfseries0.8635$\pm$0.040& 0.8505$\pm$0.036& 0.8501$\pm$0.034& 0.8505$\pm$0.039\\
48 & 0.9844$\pm$0.019& 0.9826$\pm$0.020& 0.9817$\pm$0.019& \bfseries0.9879$\pm$0.017\\
49 & \bfseries0.9707$\pm$0.041& 0.9466$\pm$0.045& 0.9411$\pm$0.051& 0.9462$\pm$0.047\\
\hline
Average & \bfseries0.8595$\pm$0.093& 0.8463$\pm$0.097& 0.8425$\pm$0.097& 0.8243$\pm$0.103\\
\hline 
\end{tabular}
\end{table}

As a first impression, looking at the tables we can observe that {\sf ODTE} and {\sf ODTE\textsuperscript{T}} are the winner algorithms both in number of wins and average accuracy. The difference is impressive in the scenario involving hyperparameter tuning, although all the algorithms are clearly benefited from the tuning process. 
To properly analyze the results, we conducted a standard machine learning statistical analysis procedure \citep{demsar_statistical_2006,garcia_extension_2008} using the {\sf exreport} tool \citep{arias_exreport_2015}. First, a Friedman test ($\alpha=0.05$) is performed to decide whether all algorithms perform equivalently. If this hypothesis is rejected, a post hoc test is performed using Holm's procedure ($\alpha=0.05$) and taking as control the highest-ranked algorithm from the Friedman test. Specifically, we have conducted two statistical analyses, one for each scenario (Tables \ref{tab:tests} and \ref{tab:tests@}).


\begin{table}[htbp]
\centering
\caption{Results of the post-hoc test for the mean accuracy of the algorithms.}\label{tab:tests}
\begin{tabular}{lrrrrr}
\hline
classifier & pvalue & rank & win & tie & loss\\
\hline
Odte & - & 3,88 & - & - & - \\
TBRRoF & \bf 1,0000e+00 & 3,95 & 27 & 0 & 22\\
TBRaF & \bf 1,0000e+00 & 4,19 & 27 & 0 & 22\\
RandomForest & \bf 8,5429e-01 & 4,47 & 27 & 0 & 22\\
BaggingJ48-SVMODT & \bf 4,8534e-01 & 4,73 & 31 & 0 & 18\\
BaggingWodt & \bf 8,2523e-02 & 5,20 & 32 & 0 & 17\\
XGBoost &  3,8055e-02 & 5,39 & 32 & 0 & 17\\
PBC4cip &  1,7426e-02 & 5,55 & 30 & 0 & 19\\
TBRoF &  9,1628e-11 & 7,63 & 45 & 0 & 4\\
\hline 
\end{tabular}
\end{table}


\begin{table}[htbp]
\centering
\caption{Results of the post-hoc test for the mean accuracy of the algorithms with optimized hyperparameters.}\label{tab:tests@}
\begin{tabular}{lrrrrr}
\hline
classifier & pvalue & rank & win & tie & loss\\
\hline
ODTE\textsuperscript{T} & - & 1,35 & - & - & - \\
TBRRoF\textsuperscript{T} &  7,6660e-04 & 2,22 & 40 & 0 & 9\\
TBRaF\textsuperscript{T} &  2,7947e-10 & 3,02 & 46 & 0 & 3\\
TBRoF\textsuperscript{T} &  7,9936e-15 & 3,41 & 44 & 0 & 5\\
\hline 
\end{tabular}
\end{table}

From the analyses we can draw the following conclusions:

\begin{itemize}
    \item In the default setting, Friedman test reports a p-value of 1,3061930e-11, thus  rejecting the null hypothesis that all the algorithms are equivalent. The results of the post hoc Holm's tests are shown in Table \ref{tab:tests} by using ODTE as control, as it is the top ranked algorithms. The column \textit{rank} represents the ranking of the algorithm obtained by the Friedman test and the \textit{p}-value represents the adjusted p-value obtained by Holm's procedure. The columns \textit{win}, \textit{tie} and \textit{loss} contain the number of times that the control algorithm wins, ties and loses with respect to the row-wise algorithm. The \textit{p}-values for the non-rejected null hypothesis are boldfaced. As can be observed, there is no statistically significant difference between ODTE and most of the algorithms, despite ODTE having a larger number of wins. The only algorithms that are clearly outperformed are {\sf XGBoost} and {\sf TBRoF}.
    
    \item In the hyperparameter tuning setting, Friedman test reports a p-value of 7,4815371e-16, thus  rejecting the null hypothesis that all the algorithms are equivalent. The results of the post hoc Holm's tests are shown in Table \ref{tab:tests@} by using {\sf ODTE\textsuperscript{T}} as control, as it is the top ranked algorithms. As can be observed, in this case there is statistically significant difference bewteen {\sf ODTE\textsuperscript{T}} and the rest of studied algorithms. 
\end{itemize}

To complete our analysis we also studied the complexity (size) of the obtained trees and also the training time required by the different oblique decision tree algorithms. For the sake of brevity\footnote{Detailed values fore each algorithm and dataset can be observed in supplementary material: {\tt https://t.ly/k1MGF} (size) and {\tt https://t.ly/WCkxm} (time)} in Table \ref{tab:size} we only show the values on average over the 49 datasets once we normalized them by using ODTE as control. 

\begin{table}[ht]
\centering
\renewcommand{\arraystretch}{1.2}
\renewcommand{\tabcolsep}{0.07cm}
\caption{Averaged normalized tree size for each dataset and algorithm.}
\label{tab:size}
\begin{tabular}{lccccccccc}\hline
&ODTE	&TBRRoF	&TBRoF	&TBRaF		&BaggJ48SVMODT&	BaggWodt & PBC4cip\\
\hline
Size & 1,00	&3,87&	2,77&	2,82&	2,66&	11,73 & --\\

Time & 1,00 &287,51&	2,02&	230,84&	10,11&	1,25 & 2,03 \\ \hline
\end{tabular}
\end{table}

With respect to {\em size} we compute the average size (number of nodes) in the trees included in the ensemble\footnote{This data could not be included for PBC4cip because the package used did not provide the average size of the trees in the ensemble.}, then we normalize by using ODTE. As all the numbers in the row are greater than one, it is clear that ODTE obtains the more compact trees. On the other hand, with respect to {\em CPU time} we must be careful with the analysis because the tested algorithms are in different programming languages. However, from the data in Table \ref{tab:size} we could obtain a similar conclusion, with ODTE being the faster algorithm.  

In summary, considering all the evaluated dimensions (accuracy, size, and time), there is no doubt that ODTE emerges as the outstanding algorithm in the comparison, a fact that is even more pronounced when we take the hyperparameter tuning phase into account.



\section{Conclusion}
\label{sec:conclu}

A new ensemble algorithm, ODTE, which incorporates a novel oblique decision tree model (STree) capable of directly handling a multi-class target variable, has been introduced. The STree algorithm generates a binary oblique decision tree that learns several SVM at each split, though only one is retained for inference. The experiments demonstrate that this approach performs well across a broad range of domains (49 datasets), surpassing competing ensembles of oblique decision trees and other state-of-the-art algorithms. 
It has also been observed that tuning the hyperparameters of the base algorithm and the ensemble for each dataset is crucial for achieving even better results, leading to a version of ODTE that significantly outperforms all other tested methods.

A limitation of our approach falls in the dimension of interpretability. Clearly, including an SVM in the decision nodes undeniably diminishes the interpretability of the resultant tree. Consequently, we believe that the dimension of explainability within the proposed STree model, akin to others emphasizing accuracy over transparency, necessitates attention through local and post-hoc explanation techniques. Obviously, this limitation of our base model (STree) extends to the proposed ensemble, ODTE, bringing it closer to the black-box family of classifiers. Fortunately, research in explainable AI (XAI) offers various techniques to shed light on the decisions made for specific instances \citep{Ribeiro:2016, Lundberg:2017, LORE2:2024}.

For future research, we plan to go deeper in the use of (random/informed) sub-spaces, selecting features randomly as in random forest \citep{Breiman:RF} or using univariate or multivariate filter feature selection algorithms. Additionally, the benefits of fine-tuning hyperparameters have proven essential for the performance of the proposed algorithms, albeit at the cost of high computational CPU time. In future studies, we aim to investigate a type of lightweight auto-tuning integrated into the ODTE/STree algorithm(s).


\section*{Acknowledgements}

First of all, we are indebted to the authors of  \cite{Ganaie:ESWA2020}, \cite{Menkovski:2008} and \cite{WODT:2019} for providing us with the code for their implementations. We are also grateful to the reviewers for their valuable comments.
This work has been partially funded by the Government of Castilla-La Mancha and ``ERDF A way of making Europe'' under project SBPLY/21/180225/000062; by MCIU/AEI/10.13039/501100011033 and ``ESF Investing your future'' through project PID2022-139293NB-C32; and by the Universidad de Castilla-La Mancha and ``ERDF A Way of Making Europe'' under project 2022-GRIN-34437.


\section*{CRediT author statement}

\noindent {\bf Ricardo Montañana}: Data curation, Software, Visualization, Validation, Conceptualization, Methodology, Writing- Original draft preparation, Writing- Reviewing and Editing. {\bf José A. Gámez, José M. Puerta}: Supervision, Funding acquisition, Conceptualization, Validation, Writing- Original draft preparation, Writing- Reviewing and Editing.

\section*{Declaration of generative AI and AI-assisted technologies in the writing process}

During the preparation of this work, the author(s) used CHATGPT and GRAMMARLY in order to improve language and readability. After using these tools/services, the author(s) reviewed and edited the content as needed and take(s) full responsibility for the content of the publication.







\end{document}